\documentclass[conference]{IEEEtran}
\IEEEoverridecommandlockouts

\usepackage{cite}
\usepackage{amsmath,amssymb,amsfonts}
\usepackage{algorithmic}
\usepackage{graphicx}
\usepackage{textcomp}
\usepackage{xcolor}
\usepackage{subcaption}
\usepackage{adjustbox}
\usepackage{booktabs} 
\usepackage{todonotes}
\usepackage{float}
\usepackage{dblfloatfix}
\usepackage{multirow} 
\usepackage{tabularx}
\usepackage{array}
\def\BibTeX{{\rm B\kern-.05em{\sc i\kern-.025em b}\kern-.08em
    T\kern-.1667em\lower.7ex\hbox{E}\kern-.125emX}}
\begin{document}

\title{GLIDE: A Coordinated Aerial-Ground Framework for Search and Rescue in Unknown Environments}

\author{\IEEEauthorblockN{Seth Farrell}
\IEEEauthorblockA{\textit{Computer Science \& Engineering} \\
\textit{University of California San Diego}\\
La Jolla, CA, USA \\
0009-0008-1497-0434}
\and
\IEEEauthorblockN{Chenghao Li}
\IEEEauthorblockA{\textit{Computer Science \& Engineering} \\
\textit{University of California San Diego}\\
La Jolla, CA, USA \\
0009-0003-8810-1710}
\and
\IEEEauthorblockN{Henrik Christensen}
\IEEEauthorblockA{\textit{Computer Science \& Engineering} \\
\textit{University of California San Diego}\\
La Jolla, CA, USA \\
0000-0002-7465-7502}
}

\maketitle


\begin{abstract}
We present a cooperative aerial-ground search-and-rescue (SAR) framework that pairs two unmanned aerial vehicles (UAVs) with an unmanned ground vehicle (UGV) to achieve rapid victim localization and obstacle-aware navigation in unknown environments. We dub this framework Guided Long-horizon Integrated Drone Escort (GLIDE), highlighting the UGV's reliance on UAV guidance for long-horizon planning. In our framework, a goal-searching UAV executes real-time onboard victim detection and georeferencing to nominate goals for the ground platform, while a terrain-scouting UAV flies ahead of the UGV's planned route to provide mid-level traversability updates. The UGV fuses aerial cues with local sensing to perform time-efficient A* planning and continuous replanning as information arrives. Additionally, we present a hardware demonstration (using a GEM e6 golf cart as the UGV and two X500 UAVs) to evaluate end-to-end SAR mission performance and include simulation ablations to assess the planning stack in isolation from detection. Empirical results demonstrate that explicit role separation across UAVs, coupled with terrain scouting and guided planning, improves reach time and navigation safety in time-critical SAR missions.
\end{abstract}

\begin{IEEEkeywords}
Multi-Agent, Robotics, SAR, Perception.
\end{IEEEkeywords}

\section{Introduction}
\label{sec:intro}
Search and rescue (SAR) operations stand to benefit from recent advances in autonomous aerial and ground robotics. Unmanned Aerial Vehicles (UAVs) enable rapid, large-area coverage due to their agility and mobility.
The adoption of drones across civilian and military applications has highlighted advantages in speed and perspective.
In SAR, aerial platforms facilitate rapid victim localization and provide global views of disaster zones.
They enable first responders to identify both victims and safe access routes,
thereby reducing response time.
Unmanned Ground Vehicles (UGVs) offer persistence, greater payload capacity, and the ability to physically reach and assist victims.
However, realizing these complementary strengths in a single operational system remains challenging due to limited UAV endurance, 
local myopic planning on UGVs,
and robust online coordination in unstructured, time-critical environments~\cite{QUERO2025105199, chen2025uavugv}.

\begin{figure}[t]
    \centering
    \adjustbox{width=0.47\textwidth}{
        \includegraphics{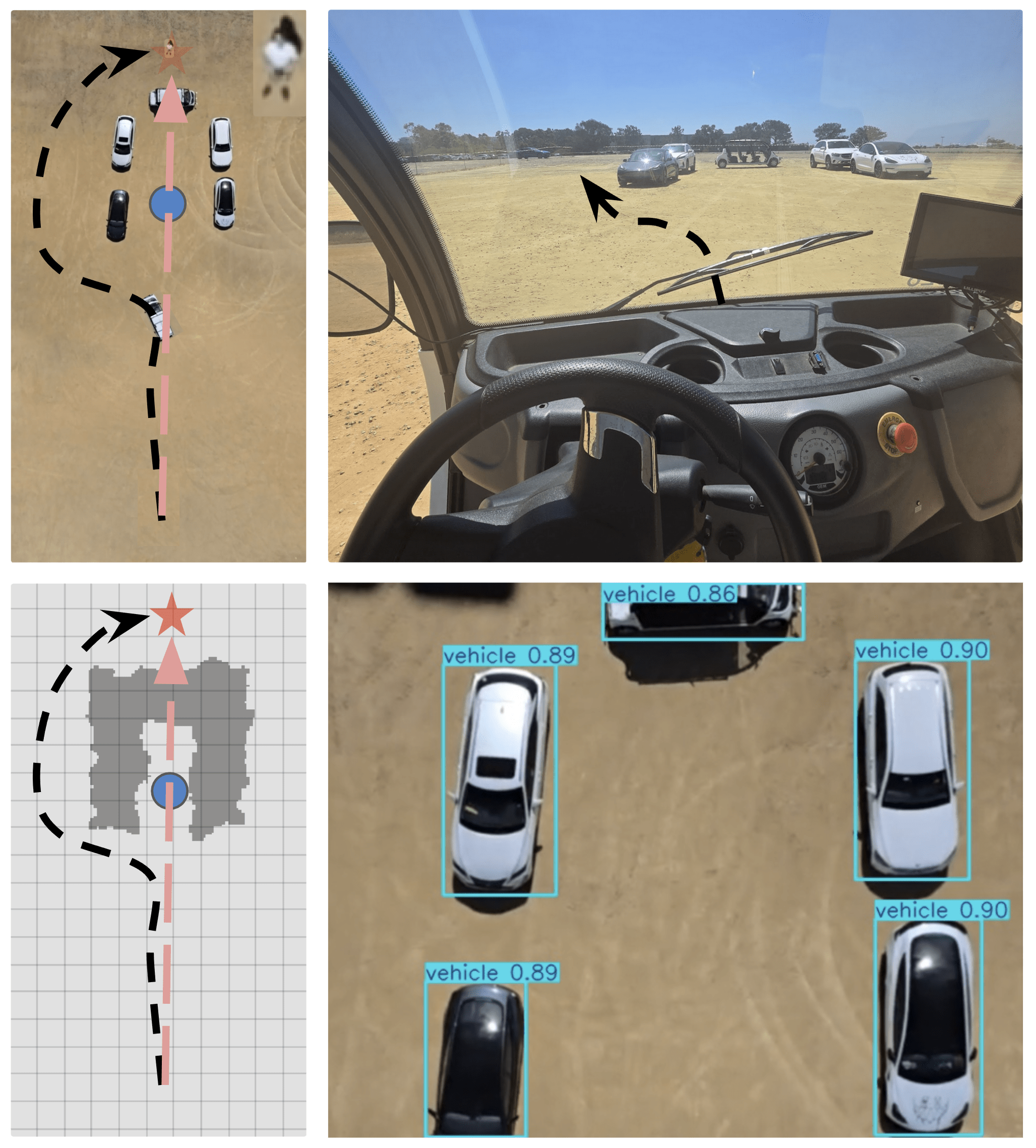}
    }
    \caption{\small 
    (Top Left) 
    Real-world experimental visualization of the proposed end-to-end aerial-ground SAR framework.
    The test environment includes a U-shaped corridor at the center,
    with the victim located at the top.
    The dashed pink line indicates the UGV's planned trajectory before receiving terrain information from the terrain-scouting UAV, while the blue dot marks the footprint of the terrain-scouting UAV.
    The black dashed line indicates the ego-vehicle's planned trajectory after re-planning occurs.
    (Top Right) Ego-vehicle perspective as it approaches the U-shaped obstacles.
    (Bottom Left) Occupancy grid generated by our GLIDE method, smoothed over time, where dark gray regions denote detected non-traversable areas.
    The ego vehicle uses this map to plan avoidance maneuvers.
    (Bottom Right) Object detection results from aerial imagery captured by the terrain-scouting UAV. The detection module outputs bounding boxes for identified objects along with their associated confidence scores.
    }
    \label{fig:main_image}
    \vspace*{-15pt}
\end{figure}

Prior SAR deployments have heavily relied on teleoperation~\cite{rs15133266},
wherein human operators must continuously monitor video feeds and sensor data, make decisions, and directly control the robots in real time.
In disaster environments, this quickly becomes overwhelming, particularly when multiple robots must be managed simultaneously. 
Moreover, teleoperation requires a reliable, high-bandwidth communication link between robots and the operator control center.
In SAR scenarios (e.g., collapsed buildings, tunnels, wildfire zones),
wireless signals are often obstructed or delayed.
Latency in control commands or video feedback can render precise maneuvering extremely difficult or even infeasible.
Finally, teleoperation underutilizes recent advances in learning-based perception, onboard inference, and edge autonomy, limiting both scalability and effectiveness.

Search-and-Rescue (SAR) operations are driven by two urgent imperatives: the need to rapidly localize victims and the necessity of delivering immediate support in unstructured, partially known environments such as uneven terrain, forests, or flooded streets. However, achieving these objectives is highly challenging due to the inherent trade-offs of different robotic platforms. UAVs, despite their speed and wide-area coverage, are constrained by limited flight time and payload capacity. UGVs, on the other hand, offer endurance and robustness on the ground but lack the rapid mobility and global situational awareness of aerial systems. The cooperation between UAVs and UGVs has emerged as a central research direction in robotics, driven by their complementary strengths \cite{mondal2023optirouteheuristicassisteddeepreinforcement} \cite{chen2025uavugvcooperativetrajectoryoptimization} \cite{brotee2024optimizinguavugvcoalitionoperations}. This recognition has inspired numerous research efforts and competitions designed to evaluate and extend the capabilities of UAV-UGV teams in realistic settings. 

We address these challenges with a cooperative multi-agent framework that integrates two aerial roles with a ground ego-vehicle for end-to-end autonomous SAR.
A \textit{goal-searching} UAV conducts fast site surveys,
detects potential victims onboard,
and transmits compact georeferenced messages to the ground platform.
A \textit{terrain-scouting} UAV scouts ahead of the UGV's current plan,
providing mid-level situational awareness by flagging non-traversable or hazardous regions.
The UGV fuses these aerial cues to perform long-horizon planning and online re-planning toward locations of confirmed victims.
This division of responsibilities accelerates victim discovery while enabling safe and time-efficient routing.
The key contributions of this paper are summarized as follows:
\begin{itemize}
    \item A cooperative aerial-ground SAR framework that explicitly separates goal-searching and terrain-scouting aerial roles to complement the ground platform. 
    \item A lightweight perception stack for real-time detection on resource-constrained UAVs.
    \item An aerial-guided planning stack that uses situational maps (from the terrain-scouting UAV) to generate time-efficient, collision-free UGV trajectories.
    \item A hardware demonstration using a GEM e6 UGV and two X500 UAVs that provides a proof of concept for end-to-end SAR automation.
    \item Simulation ablations that robustly evaluate the proposed planning stack in isolation from perception. 
\end{itemize}

\section{Related Works}
\label{sec:related}

A landmark initiative in advancing search-and-rescue (SAR) robotics was the \textit{DARPA Subterranean (SubT) Challenge}, in which participants developed robotic platforms capable of navigating underground environments and detecting designated objects. Similar to how the DARPA Urban Grand Challenge accelerated self-driving research, the SubT Challenge spurred significant progress in SAR robotics. Key outcomes included advances in UAV control within confined spaces, edge-deployable detection models, and both homogeneous and heterogeneous multi-agent planning approaches~\cite{Petrl_k_2025,depetrillo2021searchplanninguavugvteam}.

Following the SubT Challenge, research on combined UGV--UAV systems has expanded considerably. Survey works~\cite{liu2022review,shahar2025ugvuav,munasinghe2024comprehensive} highlight the complementary strengths of UAVs and UGVs, reviewing progress in communication, perception, and coordination. These studies also identify persistent challenges such as autonomy in unstructured environments, interoperability across platforms, and adaptation to dynamic operational conditions.

Building on these insights, several frameworks have been proposed to operationalize UAV--UGV cooperation. Wang et al.~\cite{wang2020collaborative} introduced an aerial--ground exploration strategy where UAVs scout unexplored regions to accelerate UGV mapping, significantly reducing mission time. Gawel et al.~\cite{gawel2018aerial} demonstrated that UAVs providing third-person perspectives can improve UGV teleoperation, underscoring the benefits of complementary sensing viewpoints. Oh et al.~\cite{oh2021smart} examined SAR scenarios under degraded communication and GPS-denied conditions. They proposed a genetic search algorithm to enable coordination when no control-station link is available, though their approach requires survivors to carry transmission devices.

More recent works extend SAR autonomy beyond UAV-focused solutions. Zafar et al.~\cite{zafar2024enhancing} employed a quadruped robot equipped with a fine-tuned YOLOv8 detector, coupled with gesture-based control, to improve victim detection and interaction in disaster settings. Other efforts~\cite{9238313} highlight the limitations of assuming static environments, proposing UAV--UGV cooperative path planning that integrates aerial image segmentation to identify safe paths in dynamic, evolving settings.

Collectively, these studies emphasize the importance of autonomy and cooperation in SAR, particularly for rapid victim localization and safe navigation. The integration of UAV--UGV frameworks stands out as a promising direction, leveraging the speed, coverage, and complementary capabilities of aerial and ground systems~\cite{liu2022review}.

\section{Guided Long-Horizon Integrated Drone Escort}
\label{sec:method}
We propose a cooperative multi-agent framework for integrated aerial-ground search-and-rescue (SAR) operations. 
Prior to detailing the framework, 
we identify the core capabilities that must be established to ensure effective SAR:
\begin{enumerate}
    \item Accurate detection and geolocation of victims.  
    \item Time-efficient path planning to reach confirmed goals. 
    \item Online situational awareness in unknown environments for long-horizon planning.
\end{enumerate} 

\begin{figure*}[t]
    \centering
    \adjustbox{width=\textwidth}{
        \includegraphics{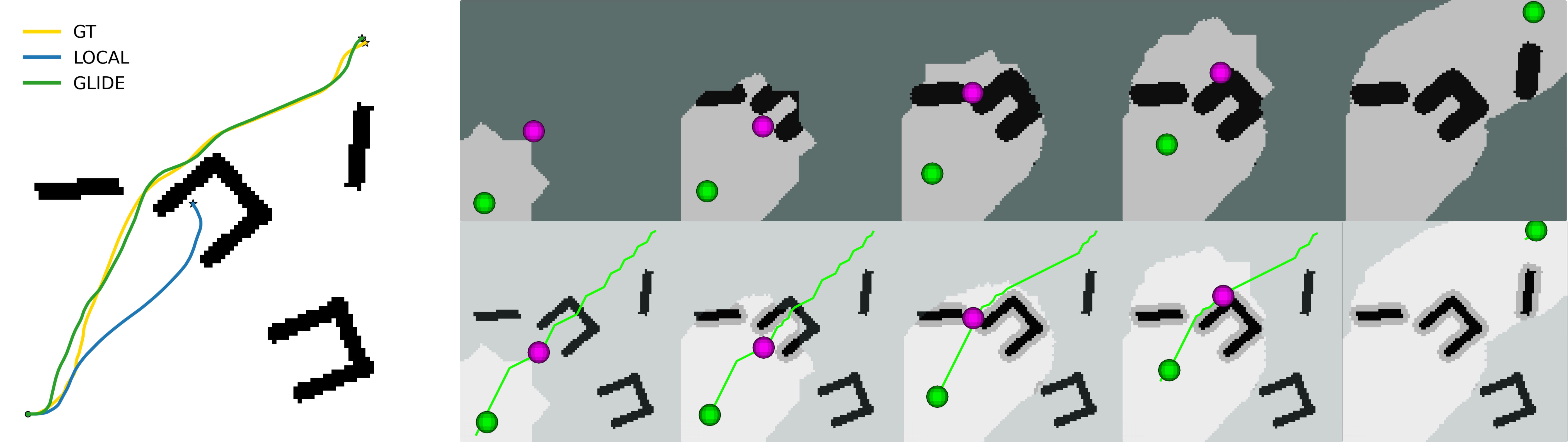}
    }
    \caption{\small
    Aerial-ground navigation in simulation.
    \textit{Left}: Trajectories under \textit{GT} (yellow), \textit{Local} (blue), and \textit{GLIDE} (green).
    \textit{Right}: Sequential snapshots of one \textit{GLIDE} trial, where the UGV (green) and UAV (purple) scouting progressively reveal obstacles (black) from the unexplored (dark grey), enabling replanning around them to reach the goal.
    }
    \label{fig:simulation_exploration}
\end{figure*}

The first two address perception and path planning, respectively,
and the third couples both: 
relying solely on a local navigation stack can cause the system to become trapped in non-convex environments (Figure \ref{fig:simulation_exploration}), 
necessitating higher-level situational awareness to obtain near-optimal solutions.

We next present a modular framework that operationalizes the three functionalities outlined above.
The system comprises a ground ego-vehicle and two complementary aerial subsystems (each instantiated by one or multiple robots) with integrated detection. 
Responsibilities are decoupled and interfaced via clearly defined information products, ensuring cohesive end-to-end behavior.
\begin{itemize}
\item \textbf{(Aerial) Goal-Searching UAV(s).}
Deployed from the ego-vehicle, 
these UAVs survey high-priority regions to detect and geolocate victims. 
\item \textbf{(Aerial) Terrain-Scouting UAV(s).}
Operating near the ego-vehicle, 
these units provide situational awareness by identifying non-traversable terrain/regions to provide a mid-level map.
\item \textbf{(Ground) Ego-Vehicle.}
The ground ego-vehicle navigates toward detected victims,
leveraging a mid-level map for long-horizon planning.
\end{itemize}

Aerial platforms offer agility and mobility that complement the ground robots,
while the ego-vehicle can be application-dependent: 
in our implementation (Section \ref{sec:implementation}), we employ a golf cart, 
but the framework is equally applicable to quadrupeds or even humanoids. 
The choice primarily affects the definition of traversability and associated constraints, 
as locomotion capabilities vary across platforms.

\subsection{Perception}
Perception underpins the proposed framework,
and we scope it into two sub-problems that provide the planning inputs: 
(i) a georeferenced set of victim coordinates, 
and (ii) a mid-level traversability map for situational awareness.

\textit{Victim Detection.}
Similar to the idea from \cite{10.1007/978-3-540-76928-6_1}, goal-searching UAVs perform victim detection from aerial imagery, 
identifying if the current frame contains target victims and generating bounding boxes for pixel-space localization.
Because these agents often operate at standoff distances from the ego-vehicle over bandwidth-limited links, 
streaming raw images is impractical.
Each agent thus executes lightweight, real-time edge inference,
transmitting only compact detection messages to the ego-vehicle.

\begin{figure}[t]
    \centering
    \adjustbox{width=0.47\textwidth}{
        \includegraphics{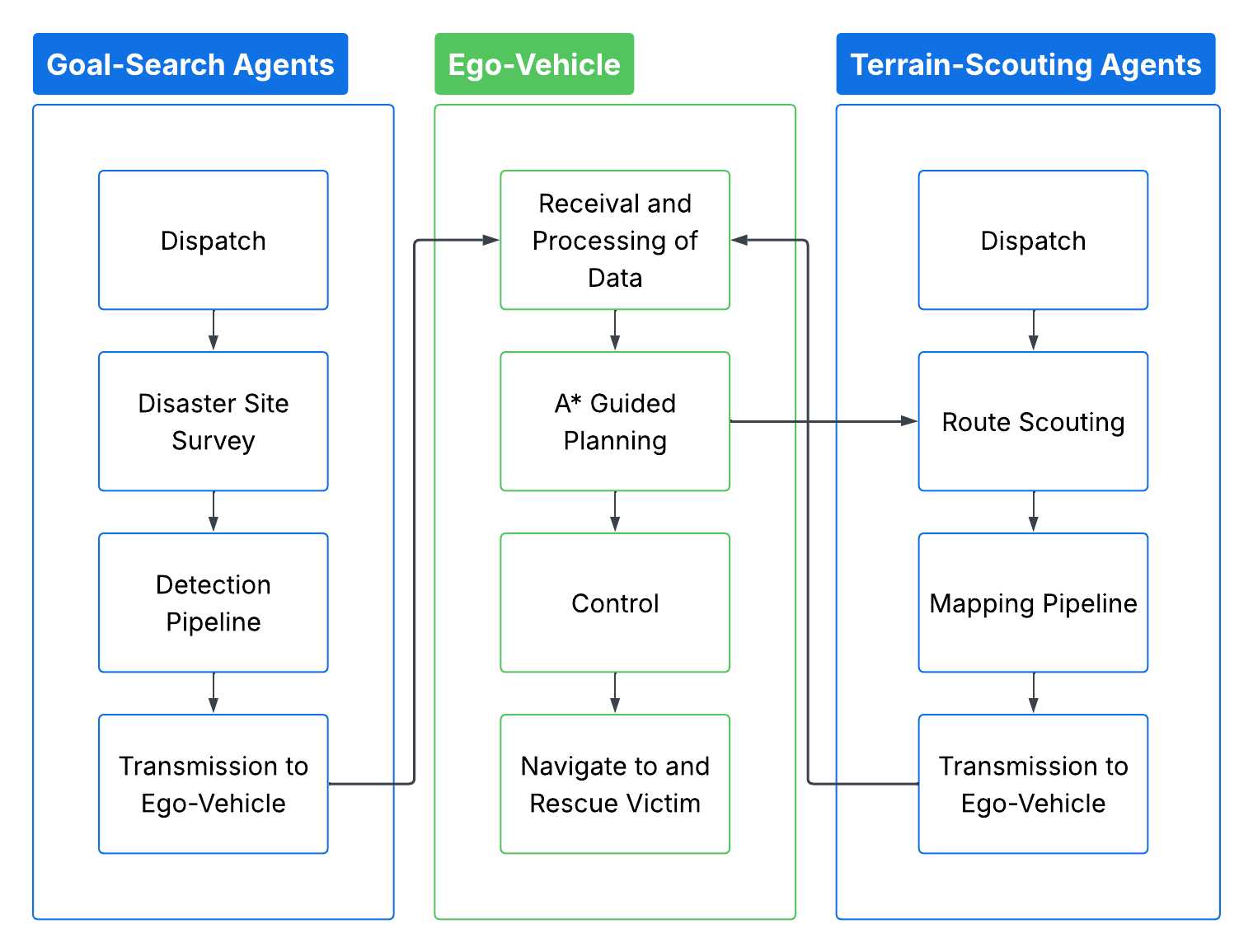}
    }
    \caption{\small The framework illustrates cooperative aerial-ground SAR operations: goal-searching UAVs survey the disaster site, detect and geolocate victims, and transmit compact messages to the UGV, while terrain-scouting UAVs track the UGV’s path to identify obstacles. The UGV fuses these inputs with local sensing and employs an A* planner to compute safe, time-optimal routes to the victims.
    }
    \label{fig:methodology_framework}
\end{figure}

\textit{Traversability inference.}
Situational awareness is derived from aerial imagery captured by terrain-scouting UAVs,
which is used to infer regions of non-traversable or hazardous terrain (e.g., water, rubble, voids). 
Since these agents operate in closer proximity to the ego-vehicle, 
transmitting compressed imagery is feasible when desired. 
Consequently, the ego-vehicle can perform the more computationally intensive tasks of segmentation and georegistration, 
while the aerial units function primarily as image capture platforms.

Since detection uses aerial images, 
all modules must be resilient to variation in altitude, viewpoint, motion blur, and illumination. 
The victim detection stack must satisfy real-time constraints on resource-limited aerial platforms,
whereas traversability inference may employ more compute-intensive methods executed on the ego-vehicle.


\begin{figure*}[t]
    \centering
    \adjustbox{width=\textwidth}{
        \includegraphics{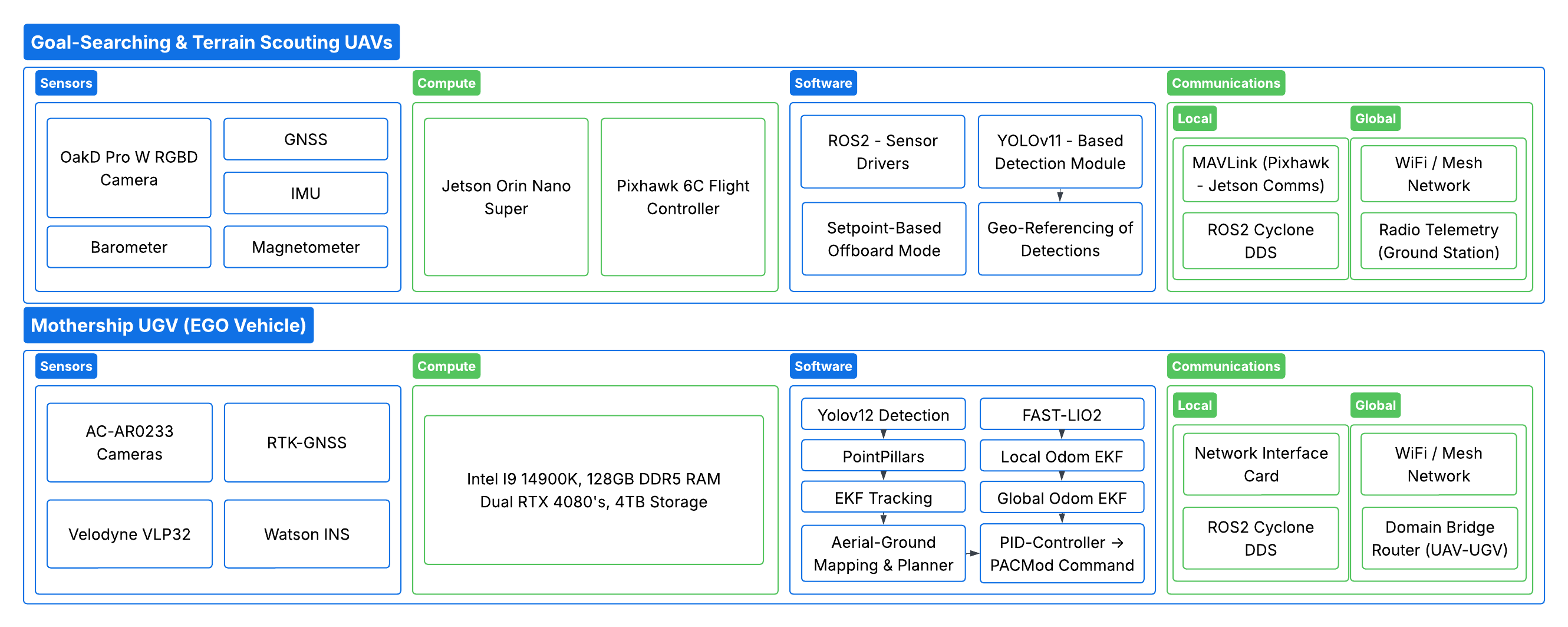}
    }
    \caption{\small Both UAVs run lightweight YOLOv11 detection pipelines, finetuned on the VisDrone dataset~\cite{zhu2021detection} and optimized with TensorRT, onboard the Jetson Orin Nano. The results are paired with the state information (retrieved from the flight controller via PyMavLink) of the UAV in order to georeference the victim positions. On the ground side, detections are ingested by the aerial mapping node, which fuses them with sensor inputs (LiDAR, RTK) to update a mid-level map. A planner node then computes safe trajectories using this map, while the control node sends low-level commands via the CAN bus to the drive-by-wire system. Together, these modules enable a coordinated, yet modular, approach to achieving the objectives of a SAR-style mission scenario.}
    \label{fig:system_flow}
\end{figure*}


\subsection{Time-Optimal Path Planning for Ego-Vehicle}
Given georeferenced victim coordinates and a mid-level traversability map, 
the ego-vehicle performs time-optimal path planning using graph-based traversal algorithms. 
To construct the planning graph, we discretize the local workspace into a uniform grid centered on the ego-vehicle,
using a predefined grid resolution parameter.

Non-traversable regions from the map are modeled as hard obstacles, 
while we insert victim coordinates as goal nodes. 
If a victim lies outside the grid,
it is projected along the corresponding direction from the ego-vehicle onto the grid boundary and treated as a proxy goal.

We formulate ego-vehicle path planning as the computation of a shortest-distance trajectory on the discretized roadmap, 
employing the graph-based search algorithm A$^{*}$.
When multiple victims are present, 
the planner constructs a pairwise shortest-time matrix over the roadmap and determines a visitation order that minimizes total travel time. 
The plan is continuously adapted, 
with replanning triggered by map updates or the detection of additional victims.

\subsection{Path Planning for Terrain-Scouting UAVs}

Terrain-scouting UAVs are assigned to follow the ego-vehicle’s planned trajectory, 
but several steps ahead. 
If the current plan has not yet reached all victims, 
a straight-line extension to the next victim in the visitation order serves as the reference. 
As unexplored regions are mapped, 
we continuously update the ego-vehicle’s path planning, 
which in turn guides the terrain-scouting UAVs toward the ego-vehicle’s anticipated future positions.

In this project, the main objective of the terrain-scouting UAVs is to validate the executability of the current planned trajectory of the ego-vehicle.
A natural extension is to incorporate advanced exploration strategies
to proactively identify alternative detours that may provide faster access to victims, which we leave as future work.

\subsection{Overall Framework}
A framework overview is shown in Figure \ref{fig:methodology_framework}. 
The operational sequence proceeds as follows. 
The system begins by deploying goal-searching UAVs for victim detection. 
Because time-optimal search planning is highly scenario-dependent and beyond the scope of this work, 
we assume that the planning paths of goal-searching UAVs are provided externally (e.g., user inputs).

Once victims are identified, 
the remainder of the system operates autonomously. 
Terrain-scouting UAVs are first dispatched ahead of the ego-vehicle in time to initialize situational awareness. 
The ego-vehicle then computes a distance-optimal trajectory to the detected victims using the progressively refined mid-level map, which encodes identified non-traversable regions. 
If the ego-vehicle becomes immobilized at any stage, 
a signal is transmitted to trigger recovery procedures.

\section{Hardware Implementation}
\label{sec:implementation}

We validate the framework with a proof-of-concept hardware demonstration.
In this section, we discuss the principal engineering choices encountered during system integration. An overview of our systems can be seen in Figure~\ref{fig:system_flow}.

\subsection{Hardware Platform}

\textbf{Ego-vehicle.}
The ground platform is a GEM e6 golf cart with actuator limits:
maximum speed $2~\mathrm{m/s}$, steering rate $40^\circ\!/\mathrm{s}$, and steering acceleration $10^\circ\!/\mathrm{s}^2$. 
Drive-by-wire control is provided by PACMOD (Platform Actuation and Control MODule) (accelerator, brake, steer, turn signals). 
Onboard compute comprises an Intel I9-14900K, 128GB RAM, dual RTX 4080 GPUs, and 4TB storage. The system is currently much more powerful than needed for GLIDE to function, allowing plenty of room for future expansion on the framework.

\textbf{Goal-Searching and Terrain Scouting UAVs.}
The UAV frames are Holybro X500 platforms (PX4 firmware) with maximum speed $5~\mathrm{m/s}$, ascent/descent rate $1~\mathrm{m/s}$, tilt limit $30^\circ$, and thrust-to-weight ratio $2$.
An outer-loop PID controller converts waypoint references to velocity commands stabilized by the onboard attitude controller.
Each airframe is modified with custom 3D-printed mounts for onboard compute and sensor integration, a DC-DC converter for stable power delivery, a USB-to-UART interface enabling DDS communication between the flight controller and companion computer, and a nadir-facing Luxonis OAK-D Pro W RGBD camera.

\begin{figure}[t]
    \centering
    \adjustbox{width=0.48\textwidth}{
        \includegraphics{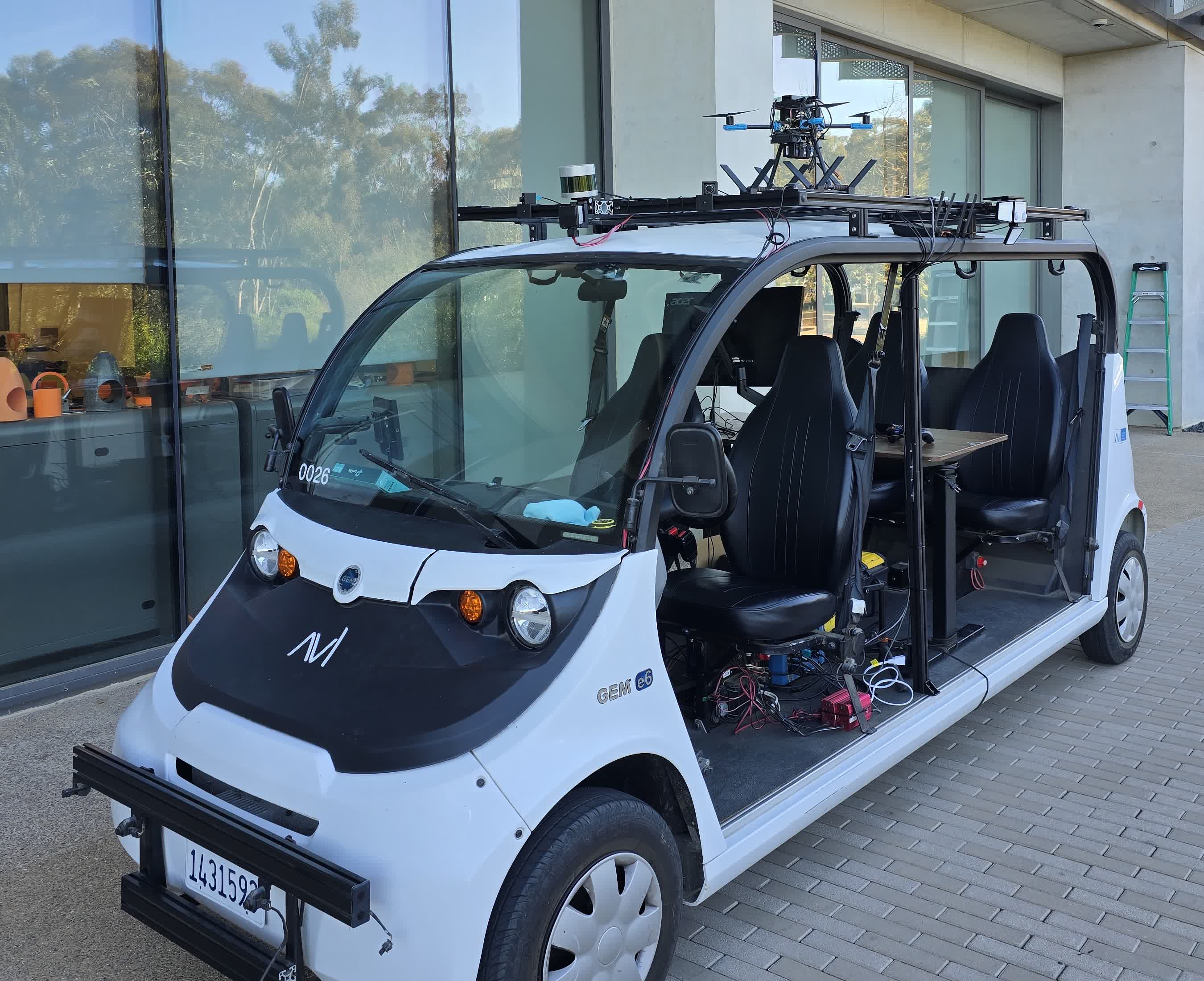}
    }
    \caption{\small The GEM e6 golf cart outfitted as the ego-vehicle for the real-world hardware demonstrations. PetG-CF 3D printed Y-shaped mounts are fixed to the top of the vehicle with 4040 aluminum extrusion to hold the drones. Aruco markers are fixed to the vehicle just under these mounts to provide precision-landing capabalities.
    }
    \label{fig:drone_and_golf_cart}
\end{figure}

\subsection{Perception}
We decouple victim and non-traversability detection into two steps:
objects are first identified in image-space,
and then mapped to the pre-defined semantic labels (e.g., people $\to$ victims; vehicles/walls $\to$ non-traversable). 

However, most commercial image-based object identification models are not directly applicable,
as they are typically trained on principal viewpoints rather than the top-down perspectives.
To this end, we fine-tune the YOLOv11~\cite{Jocher_Ultralytics_YOLO_2023} model on top-down aerial imagery from the VisDrone dataset~\cite{zhu2021detection},
with label remapping to retain only the relevant classes.

The fine-tuned model is further optimized using TensorRT and CUDA, achieving real-time inference at 30 Hz on the Jetson Orin Nano platform. The resulting lightweight computational requirements enable onboard edge inference for UAV deployment. Furthermore, the models were compressed and converted into MyriadX-compatible .blob format for direct deployment on Luxonis OAK cameras, which provide up to 4 TOPS of neural compute capability, enabling fully onboard edge detection.
In the hardware demonstrations, 
both the searching and scouting UAVs employ this YOLO-based detection stack.

A local occupancy grid used for emergency braking is generated from fused detections obtained from the Velodyne VLP-32C LiDAR and AC-AR0233 GMSL camera systems. Specifically, a PointPillars-based ~\cite{lang2019pointpillarsfastencodersobject} network is used for 3D LiDAR object detection, while YOLOv12 \cite{tian2025yolov12attentioncentricrealtimeobject} is applied across the camera streams for image-based detection. Detections from both modalities are then associated through 3D spatial consensus before being passed into a Kalman filter-based tracking pipeline. The UAV's are capable of contributing to this occupancy grid through the approached discussed in ~\ref{subsection_geolocation}.

\subsection{Localization}
Localization of the ego-vehicle is performed primarily using LiDAR-Inertial Odometry (LIO), augmented with GPS-RTK measurements with NTRIP corrections from the California Real Time Network (CRTN) system of base-stations. Specifically, FAST-LIO ~\cite{xu2021fastliofastrobustlidarinertial} is used to provide locally consistent odometry estimates, while the RTK system anchors the platform within a shared ENU global reference frame used by both the UGV and UAV systems. The Inertial-Navigation-System utilized on the platform is equipped with a dual-antenna setup which allows us to access true north heading for global orientation alignment. The local and global odometry estimates are smoothed using an EKF within the robot localization package as is standard for ROS systems. Additionally, all coordinate frames conform to the ROS REP-103 standard \cite{rep103}, providing a consistent ENU-based reference framework for localization, mapping, and multi-agent coordination. The UAV's rely on the internal state estimation of the Pixhawk flight controller operating in the North-East-Down (NED) reference frame.

\subsection{Communication}
A TP-Link Archer AX21 (AX1800) router mounted on the golf cart provides the network backhaul.
Since the internet throughput decreases with range,
only compact georeferenced messages ($<1$ kB/event) are transmitted between the UAVs and the ego-vehicle.

If connectivity is lost, detections are queued onboard and forwarded upon reconnection.
This design ensures that, even during long-range missions,
the UAVs continue collecting information and, 
once they return to a region with reliable connectivity,
transmit the accumulated data to the ego-vehicle.

\subsection{Geolocation}
\label{subsection_geolocation}
Detections from the nadir-facing camera on the UAVs are projected to the ground plane using camera intrinsics together with GPS and altitude measurements.
Each object detection, represented by a bounding box from the fine-tuned YOLO model, 
is approximated as a ground contact in pixel space.
The drone quaternion, which encodes the camera heading, 
is then used to enable a closed-form projection from image coordinates to georeferenced world coordinates.

Accurate timestamp synchronization across image, GPS, and altitude is required.
Under stable attitudes, meter-level accuracy is typical.
GPS dominates the error budget, 
while RTK corrections yield a sub-meter localization.

\subsection{Planning}
Goal-searching UAVs provide two planning options:
(i) autonomous waypoint surveys uploaded via QGroundControl with a site geofence,
and (ii) manual pilot supervision for scenario-dependent exploration.
The perception stack (utilizing Docker Compose and pymavlink) bridges the flight controller and Jetson,
launches sensor drivers, runs the ROS2 detection node,
georeferences detections,
and transmits/caches results.

The ego-vehicle incrementally updates the roadmap to incorporate georeferenced victim detections and free-space waypoints,
while treating non-traversable regions as hard obstacles.
LiDAR returns are fused during graph construction to enforce local safety.
The planning graphs are persistent throughout the mission,
enabling the system to accumulate situational knowledge over time rather than discarding previously constructed structures.
Trajectories from the time-optimal planner are executed by a rate-limited PID tracking controller.

\section{Simulation Experiments}
\label{sec:simulation}
We developed a PX4-Gazebo (Harmonic) simulation that mirrors the proposed framework while isolating the planning stack from perception to enable controlled ablations.
The setup uses ROS2 Humble, a Clearpath Warthog UGV, and a PX4 X500 UAV.
The simulation will be open-sourced along with the entire codebase.

\textbf{Scenarios.}
Worlds are procedurally generated (SDF) to create U-shaped corridors and linear barriers with randomized sizes and orientations.
We design two levels of difficulty characterized by different object densities. In Figure~\ref{sec:simulation}, an example of an easy scenario under exploration is shown.

\textbf{Settings.}
We consider all three planning strategies presented in Section \ref{sec:realexperiments}.
The simulation assumes perfect detection:
a $50$m$\times$$50$m area centered at the terrain-scouting UAV's footprint is treated as detected,
and all non-traversable information within this area is incorporated into the ego-vehicle's map.
The $A^*$ planner is evaluated under two heuristic configurations, $A^{*}_{1}$ and $A^{*}_{2.5}$, which differ in the orientation-penalty coefficient $\lambda$ ($1.0$ vs $2.5$).
Forty trials are run per scenario and setting.

\begin{table}[h]
\centering
\begin{tabularx}{\linewidth}{
c
c
c
>{\centering\arraybackslash}X
>{\centering\arraybackslash}X
>{\centering\arraybackslash}X
}
\toprule
Difficulty & Algorithm & Setting & Duration (sec) & Mean Distance (m) & Success Rate (\%) \\
\midrule
\multirow{6}{*}{Easy} & \multirow{3}{*}{$A^{*}_{1}$} & \textit{GT} & 50.08 & 66.89 & 100 \\
                      &                     & \textit{Local} & 59.25 & 92.10 & 70 \\
                      &                     & \textit{GLIDE} & 51.28 & 74.55 & 100 \\
\cmidrule{2-6}
                      & \multirow{3}{*}{$A^{*}_{2.5}$} & \textit{GT} & 53.73 & 67.34 & 100 \\
                      &                     & \textit{Local} & 60.08 & 86.76 & 75 \\
                      &                     & \textit{GLIDE} & 56.80 & 72.99 & 100 \\
\midrule
\multirow{6}{*}{Hard} & \multirow{3}{*}{$A^{*}_{1}$} & \textit{GT} & 52.54 & 73.76 & 100 \\
                      &                     & \textit{Local} & 63.50 & 105.64 & 65 \\
                      &                     & \textit{GLIDE} & 59.37 & 84.23 & 90 \\
\cmidrule{2-6}
                      & \multirow{3}{*}{$A^{*}_{2.5}$} & \textit{GT} & 57.60 & 81.00 & 95 \\
                      &                     & \textit{Local} & 64.98 & 108.91 & 75 \\
                      &                     & \textit{GLIDE} & 59.82 & 91.43 & 95 \\
\bottomrule
\end{tabularx}
\caption{\small Results of Simulation Testing}
\label{table:simulation_results_table}
\end{table}

\textbf{Results.}
Across both difficulty levels, \textit{GLIDE} consistently narrows the gap to \textit{GT} and substantially outperforms \textit{Local} (Table \ref{table:simulation_results_table}),
consistent with the hardware results.
On easy worlds, \textit{GLIDE} achieves $100\%$ success rate for $A^{*}_{1}$/$A^{*}_{2.5}$ versus $70-75\%$ for \textit{Local},
with small overheads relative to \textit{GT} ($+1.2\text{s}/+7.7\text{m}$ for $A^{*}_{1}$; $+3.1\text{s}/+5.7\text{m}$ for $A^{*}_{2.5}$).
On hard worlds, \textit{GLIDE} improves success to $90-95\%$ (vs. $65-75\%$ for \textit{Local}) with moderate overheads relative to \textit{GT} ($+6.8\text{s}/+10.5\text{m}$ for $A^{*}_{1}$; $+2.2\text{s}/+10.4\text{m}$ for $A^{*}_{2.5}$).
Increasing the A* heuristic coefficient $\lambda$ from $1$ to $2.5$ biases $A^{*}$ toward safer but more conservative trajectories:
on hard worlds, the mean path length of \textit{GLIDE} increases by $7.2$m, while the success rate improves by $5\%$.
Relative to Table~\ref{table:hardware_results_table}, 
the \textit{GT}-\textit{GLIDE} gap is comparable in several settings despite procedurally generated clutter,
largely because simulation assumes perfect detection and georegistration.
These results motivate improving the planning of terrain-scouting UAVs to proactively find shorter detours and further reduce time-to-rescue.

\textbf{Discussion.} The simulation study provides several insights that complement the hardware experiments. 
First, the gap between \textit{GLIDE} and \textit{GT} is consistently smaller in simulation than in reality. 
This is largely because the simulation assumes perfect perception: the UAV can instantaneously and without error integrate all obstacles within a fixed detection radius. 
In real-world deployments, however, perception is inevitably limited; sensors provide only partial coverage, suffer from noise, and update maps with latency. 
As a result, some obstacles may be missed or inaccurately localized, leading to suboptimal planning and a wider gap between \textit{GLIDE} and \textit{GT}. 
These findings emphasize the need for perception-aware planning strategies that are robust to incomplete or uncertain environmental information.

\section{Hardware Experiments}
\label{sec:realexperiments}


We validate the framework in outdoor trials aimed at searching for victims and driving the UGV to their georeferenced locations while minimizing travel time and ensuring collision-free navigation. A system overview of the framework discussed is shown in Figure ~\ref{fig:system_flow}.

\textbf{Scenarios.} 
Owing to limited hardware resources (primarily objects to construct non-traversable areas), 
we restrict our evaluation to two scenarios: 
(i) \textit{U-shaped corridor} formed by parked vehicles (opening facing North) and (ii) \textit{Linear barrier} (East-West), 
both with the UGV starting North and the victim South (Figure \ref{fig:main_image}).

\textbf{Settings.}
We compare three planning strategies where the system plans with:
(i) a ground-truth map (\textit{GT}),
(ii) only onboard LiDAR with a $10$m$\times 10$m local window (\textit{Local}),
and (iii) the traversability map supplied by the terrain-scouting UAVs that fly ahead of the current UGV plan with a $15$m lead offset (\textit{GLIDE}).
In all trials, the goal-searching UAV first localizes the victim and transmits the georeferenced detection.
We apply $A^{*}$ on a $300\times300$ grid ($0.5$m resolution in ENU coordinates),
using a distance-plus-orientation heuristic that favors states both spatially close to the goal and properly aligned in heading:
\begin{align*}
    h(s, g) = \|(x, y) - g\|_{2} + \lambda \cdot \Delta \theta / \pi, \label{eq:A_heuristics}
\end{align*}
where $s = (x, y, \Delta\theta)$ is the state, $g$ is the goal, $(x, y)$ is the position,
$\Delta \theta$ is the minimal heading difference, and $\lambda$ is the orientation-penalty coefficient.
For hardware experiments, we set $\lambda = 1$. 
To mitigate georegistration errors,
we discard detections when UAV roll or pitch exceeds $15^\circ$.
We fly at $15$m altitude, admit detections only from the image center,
and require a $3$-hit spatial consensus within $1$m before insertion into the map.
Ten runs are executed per scenario and setting with start poses randomized by $\pm5$m.
Trials terminate upon goal reach or safety-driver intervention.

\begin{table}[H]
\centering
\begin{tabularx}{\linewidth}{
c
c
c
>{\centering\arraybackslash}X
>{\centering\arraybackslash}X
>{\centering\arraybackslash}X
}
\toprule
Scenario & Setting & Duration (sec) & Mean Distance (m) & Success Rate (\%) \\
\midrule
\multirow{3}{*}{U-Shaped} & \textit{GT} & 43.9 & 102.40 & 100 \\
                          & \textit{Local} & 51.27 & 117.8 & 20 \\
                          & \textit{GLIDE} & 47.17 & 106.20 & 100 \\
\midrule
\multirow{3}{*}{Line} & \textit{GT} & 44.70 & 87.59 & 100 \\
                      & \textit{Local} & 56.70 & 146.45 & 60 \\
                      & \textit{GLIDE} & 50.29 & 108.73 & 100 \\
\bottomrule
\end{tabularx}
\caption{\small Results of Hardware Demonstration}
\label{table:hardware_results_table}
\end{table}

\textbf{Results.}  Table \ref{table:hardware_results_table} summarizes the averaged metrics.
The \textit{GLIDE} setting achieves $100\%$ success in both scenarios and greatly surpasses \textit{Local} in the success rate,
highlighting the importance of long-horizon planning guided by situational awareness.
Relative to \textit{GT}, \textit{GLIDE} remains close in the U-shaped case ($+3.27$s, $+3.7\%$ path) and exhibits a conservative detour in the Line case ($+5.59$s, $+24.1\%$ path).
Such discrepancy is driven by whether the terrain scouts spent enough time in the area at stable orientations to update the map. Additionally, due to the U-shaped structure, the planned path takes on a much larger change after discovering the whole obstacle compared to a simple line.
The results reflect that further enhancements require advances in UAV task planning to enhance traversability mapping.
Overall, we demonstrate that the proposed framework enables search and rescue in unknown environments, using UAVs to guide UGV planning towards the victims.

\textbf{Discussion.} 
From our hardware experiments, we uncovered several key insights about our GLIDE framework. For example, the large performance gap between the \textit{Local} and our \textit{GLIDE} setting underscores the severe limitation of navigating under a purely reactive navigation strategy, especially when facing non-convex environments. Without the aerial guidance provided by GLIDE, the ego-vehicle becomes trapped in local minima because it is unable to perceive the global structure for effective path planning.

Within the U-shaped scenario, we note that \textit{GLIDE} achieves excellent performance, suggesting that the terrain-scouting UAV captures the essential geometry of the obstacle. However, in the line-shaped scenario, there is a much larger performance gap between \textit{GLIDE} and the \textit{GT} setting. This was unexpected, as we presumed that the simpler scenario should prove less challenging. We surmise that due to the less complex shape, the terrain-scouting UAVs spend less time hovering over the obstacles, resulting in sparser coverage and more conservative path planning by the ego-vehicle.

Some practical considerations also emerged during our field trials. First, ensuring stable UAV orientations during perception tasks was critical for accurate georeferencing. Without a gimbal attachment, we improvised by enforcing a pitch/roll threshold when performing the perception tasks. Additionally, the 3-hit consensus methods helped to filter incorrect mappings that escaped the previous thresholding approach. Motion blur was not an issue in our case due to the use of a global shutter camera, though the filtering approaches mentioned above would help in that case as well. Additionally, wind gusts, lighting conditions, and the terrain texture itself occasionally degraded performance, suggesting a direction for future work to improve upon it.


\section{SUMMARY}
\label{sec:summary}

In this paper, we propose and validate a cooperative aerial-ground search-and-rescue (SAR) framework that pairs two UAVs with a UGV to enable rapid victim localization, reduce time-to-rescue, and improve navigational safety. Our system assigns complementary aerial roles to the UAVs: goal searching and terrain scouting. The UAVs employ lightweight, fine-tuned YOLO-based detectors to identify and georeference victims as well as flag untraversable terrain, transmitting this information to the UGV to support planning. Our results show that explicitly decoupling aerial roles for detection and mapping, then closing the loop with UGV planning, yields reliable and fast mission completion in SAR scenarios. These results are verified in both simulation and the real world.

Future work could include incorporating online search-planning algorithms that adaptively cover high-probability victim locations to further enhance system autonomy. Additionally, the fixed 15~m lead offset for terrain-scouting UAVs could be dynamically adjusted based on UGV speed and environmental complexity to better optimize the exploration-exploitation trade-off.


\bibliographystyle{IEEEtran}
\bibliography{references}

\end{document}